%% file: main.tex
\documentclass[runningheads]{llncs}
\usepackage[T1]{fontenc}
\usepackage{graphicx}
\usepackage{booktabs}
\usepackage[misc]{ifsym}

\usepackage[fencedCode]{markdown}
\usepackage{amsmath,amssymb,amsfonts}
\usepackage{algorithmic}
\usepackage{graphicx}
\usepackage{textcomp}
\usepackage{xcolor}

\usepackage{xspace}
\usepackage{todonotes}
\usepackage{multirow}
\usepackage{mathtools}
\usepackage{graphicx}
\graphicspath{ {./content/images/} }
\usepackage[hidelinks]{hyperref}    %
\usepackage{cleveref}
\usepackage{placeins}
\usepackage[caption=false, font=footnotesize]{subfig}
\usepackage{wrapfig}
\usepackage[export]{adjustbox}
\usepackage{array}
\usepackage{comment}

\DeclarePairedDelimiter{\norm}{\lVert}{\rVert}
\newcommand{\repeatthanks}{\textsuperscript{\thefootnote}}
\includecomment{confidential}

\begin{document}

\title{Walking Noise: On Layer-Specific Robustness of Neural Architectures against Noisy Computations and Associated Characteristic Learning Dynamics}
\toctitle{Walking Noise: On Layer-Specific Robustness of Neural Architectures against Noisy Computations and Associated Characteristic Learning Dynamics}

\titlerunning{Walking Noise}

\author{Hendrik Borras\thanks{These authors share first authorship, with different emphasis on methodology, experimentation, data analysis and research narrative.}\hspace{0.5mm}\href{https://orcid.org/0000-0002-2411-2416}{\includegraphics[scale=0.08]{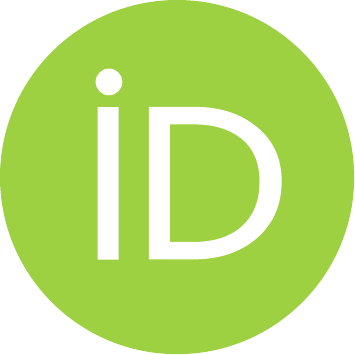}} \and
	Bernhard Klein\repeatthanks\hspace{0.5mm}\href{https://orcid.org/0000-0003-0497-5748}{\includegraphics[scale=0.08]{content/images/orcid.pdf}} \and
	Holger Fröning\hspace{0.5mm}\href{https://orcid.org/0000-0001-9562-0680}{\includegraphics[scale=0.08]{content/images/orcid.pdf}}}
\tocauthor{Hendrik Borras,Bernhard Klein,Holger Fröning}

\authorrunning{H. Borras et al.}

\institute{Computing Systems Group, Institute of Computer Engineering, Heidelberg University, Germany\\
\email{\{hendrik.borras,bernhard.klein,holger.froening\}@ziti.uni-heidelberg.de}}

\maketitle              %

\input{content/0-abstract}

\input{content/1-intro}

\input{content/2-related-work}

\input{content/3-methodology}

\input{content/4-add-noise}

\input{content/5-mul-noise}

\input{content/5b-comb-noise}

\input{content/6b-using_WN}

\input{content/6-summary}

\begin{credits}
\subsubsection{\ackname} 
This work is part of the COMET program within the K2 Center “Integrated Computational Material, Process and Product Engineering (IC-MPPE)” (Project No 886385), and supported by the Austrian Federal Ministries for Climate Action, Environment, Energy, Mobility, Innovation and Technology (BMK) and for Labour and Economy (BMAW), represented by the Austrian Research Promotion Agency (FFG), and the federal states of Styria, Upper Austria and Tyrol.

The authors additionally acknowledge support by the state of Baden-Württemberg through bwHPC and the German Research Foundation (DFG) through grant INST 35/1597-1 FUGG.\\

\end{credits}

\bibliographystyle{splncs04}
\bibliography{references}

\begin{confidential}
\clearpage
\input{content/8-Appendix}

\end{confidential}

\end{document}

%% file: content/0-abstract.tex
\begin{abstract}
Deep neural networks are extremely successful in various applications, however they exhibit high computational demands and energy consumption. This is exacerbated by stuttering technology scaling, prompting the need for novel approaches to handle increasingly complex neural architectures.
At the same time, alternative computing technologies such as analog computing, which promise groundbreaking improvements in energy efficiency, are inevitably fraught with noise and inaccurate calculations.
Such noisy computations are more energy efficient, and, given a fixed power budget, also more time efficient.
However, like any kind of unsafe optimization, they require countermeasures to ensure functionally correct results.

This work considers noisy computations in an abstract form, and gears to understand the implications of such noise on the accuracy of neural network classifiers as an exemplary workload.
We propose a methodology called \emph{Walking Noise} which injects layer-specific noise
to measure the robustness and to provide insights on the learning dynamics. 
In more detail, we investigate the implications of additive, multiplicative and mixed noise for different classification tasks and model architectures.
While noisy training significantly increases robustness for all noise types, we observe in particular that it results in increased weight magnitudes and thus inherently improves the signal-to-noise ratio for additive noise injection.
Contrarily, training with multiplicative noise can lead to a form of self-binarization of the model parameters, leading to extreme robustness.
We conclude with a discussion of the use of this methodology in practice, among others, discussing its use for tailored multi-execution in noisy environments.
\keywords{noisy training  \and noisy computations \and analog computing \and robustness \and neural networks.}
\end{abstract}

%% file: content/1-intro.tex
\section{Introduction}
\label{sec:intro}

Over the last two decades the computational demands of modern machine learning (ML) solutions have increased continuously at an exponential scale~\cite{AiCompute}.
In contrast, advances in digital technology, such as new process nodes, cannot keep up by orders of magnitude~\cite{DBLP:journals/corr/abs-2007-05558},
therefore, new architectures and technologies are coming under close consideration.

In this regard, multiple recent works have considered \textit{noisy hardware} to improve energy efficiency and thus performance, for instance based on analog electrical computations~\cite{9197673,1564344,Schemmel2022}, analog optical computations~\cite{Lin2018,Shen2017} and similarly emerging memory technologies such as resistive RAM~\cite{9127144}.
Considering the example of analog electrical systems, computations can be represented using the physical laws of network analysis, such that for instance
the input operands of a Multiply Accumulate (MAC) operation are current pulses, with the pulse's length and amplitude representing activation and weight, respectively.
A single multiplication is then the time integral over this pulse, therefore charge.
While in theory this analog integration does not consume any energy, in practice effects like inductance, resistance and thermal noise dissipate small amounts of power.
Nevertheless, for up to 8 bits of precision, analog computations are considered superior to their digital counterpart in terms of energy efficiency~\cite{9197673}.

To use such noisy hardware in practice, methods to improve robustness against unreliable computations are essential.
For instance, \textit{Noisy Machines}~\cite{NoisyMachines} explores the use of knowledge distillation from a digitally trained teacher network to a student network with noise injection.
Similarly, training as a counter measure to improve robustness against noise while slowly exposing a neural network architecture to an increasing amount of noise has proven to be effective~\cite{item2021}.
In general \emph{noisy training} has been established as a vital training methodology for different noisy hardware architectures~\cite{joshi2020,rekhi2019,NoisyMachines}.
However, a detailed analysis and in-depth understanding of how model architectures achieve robustness is still missing, which is essential to guide the search for new techniques to increase robustness even further.

In the presented work, we are concerned with an assessment of the robustness of various model architectures on different datasets against unreliable hardware.
In detail, we propose a method to measure the implications of different types and amounts of injected noise on the prediction accuracy of the model architecture and analyze the inherent defense strategies of neural networks, when trained with noise injection.
While usually both additive and multiplicative noise types are found in almost any technology, typically one type is dominant. Furthermore, this also highly depends on the chosen technology~\cite{Schemmel2022,wu2022harnessing,10.1145/3316781.3317888}.
Thus, we first investigate both noise types individually, then their combination.
This work makes the following overarching contributions:
\begin{enumerate}
\item We propose a method called \emph{Walking Noise} that assesses the implications of noise sources on prediction accuracy, which is based on walking selectively through the architecture evaluating robustness of individual layers by means of a metric called \emph{midpoint noise level}. We show why this method is necessary, and how it allows to understand the sensitivity of different architecture components.

\item In a first step the implications of additive noise are investigated for different architectures.
As we will see, one can observe that model architectures tend to adapt their parameters to increase robustness and sustain accuracy.
\item A similar methodology is applied to multiplicative noise, highlighting observations and insights. Surprisingly, some networks learn to self-binarize, leading to extreme robustness against noise.
\item The influence of batch normalization (BN) on how networks learn to tolerate injected noise is investigated.
\item We further analyze the robustness in a mixed noise setting and observe general invariance to small perturbations and high robustness, when multiplicative noise is injected prior to additive noise.
\item We show on the example of selective multi-execution how a countermeasure to noise can be tailored.

\end{enumerate}
Finally, we will shortly highlight how insights and results from \emph{Walking Noise} can be used in practice.
Overall this work aims to highlight the importance of noise-tolerant neural networks for uncertain, analog hardware and contributes first insights to understand how neural networks learn to tolerate noise on a per-layer level.

%% file: content/2-related-work.tex
\section{Related Work}
\label{sec:related}

Since the early days of machine learning, noise injection has been considered as a generalization method~\cite{murray1994,grandvalet1997noiseinjection}.
Jiang~\textit{et al.} compared it with other methods like weight decay and early stopping to reduce overparametrization~\cite{jiang2009effectNI}.
With the advent of adversarial attacks, the search for defense strategies has become an important research topic.
Alongside adversarial training~\cite{goodfellow2015adversarialtraining}, noisy training has proven to be a very effective defense, considering for instance globally injected additive Gaussian noise~\cite{He2019parametricNoise} or ensembles with layer-wise noise injection~\cite{Liu2018selfensemble}.
Liu~\textit{et al.} investigate how the robustness of continuous neural networks~\cite{chen2018neuralODE} against adversarial attacks can be improved through noise injection --- differentiating between diffusion (Gaussian distributions) and jump term (dropout) randomness~\cite{liu2020helpRobustness}.

While most of these related works evaluate input sensitivity and noise tolerance at the network inputs, noisy hardware rather affects computations, thus in particular the inherent dot products of neural networks.
To counteract accuracy degradation due to noisy computations, noisy training has also been successfully applied:
while \cite{joshi2020} and \cite{rekhi2019} inject additive zero-mean Gaussian noise with different variances during training, \textit{Noisy Machines} \cite{NoisyMachines} extends this further with knowledge distillation.
Moreover, the latter was able to explain the higher sensitivity of deeper model architectures in comparison to wider model architectures, as a loss of information through a mutual information analysis.

In particular cases, noisy hardware has been used as an inherent advantage for the deployed neural network;
for instance noise in electrical or optical analog circuits has been used to increase robustness against adversarial attacks~\cite{cappelli2022} and to support generative adversarial networks with noisy training~\cite{wu2022harnessing}. As well as to directly exploit configurable noise to execute Bayesian Neural Networks, based on Stochastic Variational Inference~\cite{ProbLight}.

While most related works show the benefit of \emph{noisy training} for generalization, robustness and resilience against noisy analog hardware, this work aims to shed light on the internals of model architectures %
trained with noise.

%% file: content/3-methodology.tex
\section{Methods}
\label{sec:method}

Depending on the accelerator, noise effects can appear at varying locations during the inference of a neural network.
Considering that most ML accelerators are based on a matrix multiplication unit as the central computational unit, three places for noise injection become apparent:
(1) at the weight readout, 
(2) inside the calculation,
(3) when forwarding activations to subsequent layers.
In this work we focus our investigation on the last injection point,
since this point can combine effects from the first two injection points. 
Furthermore, we consider additive, multiplicative and, mixed injection of Gaussian noise, due to the distributions widespread observation in natural processes and success in other works on noise injection~\cite{NoisyMachines,joshi2020}.
To inject noise without bias, we sample with zero and unit mean, for the additive and multiplicative case, respectively.

Such noise is injected either only during inference or during training and inference.
Where the first approach corresponds to running an unmodified model on a noisy accelerator, the latter 
has been shown to improve network accuracy significantly~\cite{item2021}.

\subsection{Global and Walking Noise}
\label{sec:method:gaw_noise}

To inject noise at the activations of a given neural network, we apply a custom noise module after each layer.
With this approach we can selectively inject noise at any given point of the activation path of a network, even directly at the input and output.
Similar to straight-through estimators used in quantization-aware training~\cite{Jacob_2018_CVPR}, we only inject noise in the forward path, not the backward path.

For our initial experiments we injected noise globally with the same intensity at all layers of the network.
While these experiments already give a first impression of the sensitivity of a network to noise, no information is gained about the network's internal behavior.
To probe how the internals of a model architecture react to noise injection, we inject noise exclusively at a single layer. 
By moving the injection point between experiments, the noise is then walking selectively through the network, coining the term \emph{Walking Noise}.

\subsection{Data sets, models and experimental setup}

We consider multiple Image Classification (IC) tasks (MNIST~\cite{MNIST}, FashionMNIST~\cite{FashionMNIST}, SVHN~\cite{SVHN}, CIFAR-10~\cite{cifar10}) and one Natural Language Processing (NLP) task (Google Speech Commands (GSC) v2~\cite{speechcommandsv2}).
In the interest of conciseness, we focus on reporting CIFAR-10 results, which transfers to the other datasets if not otherwise stated.
Only GSC was preprocessed, using MFCCs with settings according to the MLPerf Tiny benchmark \cite{tinymlbench}.

For this work we consider three different model architectures: a Multi-layer Perceptron (MLP), LeNet-5~\cite{726791}, and CNN-S~\cite{helloEdge}.
The MLP consists of three fully-connected (FC) layers with 64 neurons, ReLU activations and optional batch normalization between these layers, and is used for all tasks. 
LeNet-5 is used for IC tasks %
while for GSC the CNN-S architecture is used, due to a significant difference in input shape.

\label{sec:method:exp} 
In total over $490,000$ experiments were conducted.
Confidence intervals used in the regression fits are based on multiple training runs with different random seeds.
The code for all experiments is publicly available\footnote{\url{https://github.com/UniHD-CEG/walking-noise}}.

\section{Analysis of global noise}

\label{sec:method:ana}
\begin{figure*}%
	\centering
	\includegraphics[width=0.75\columnwidth]{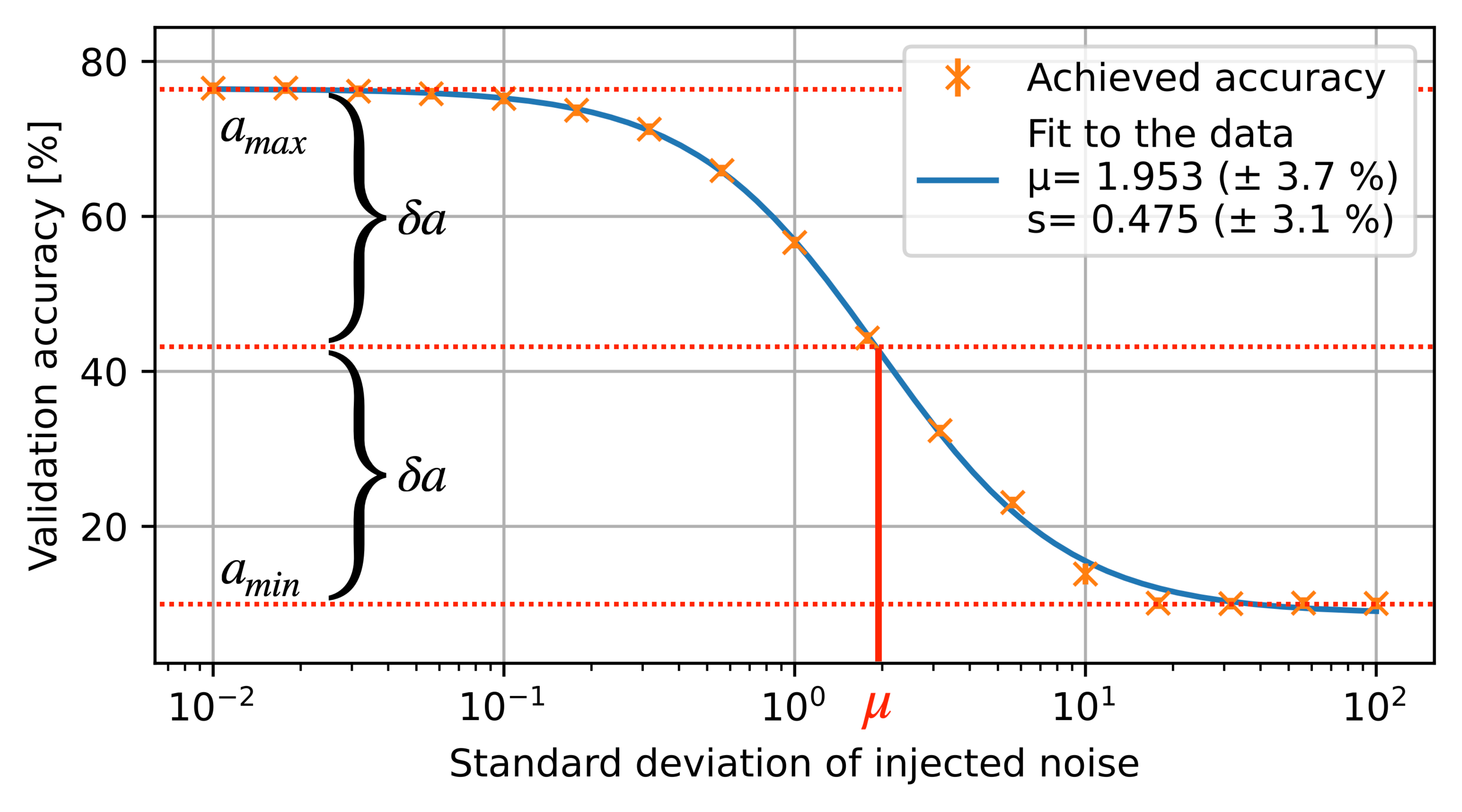}
	\vspace{-0.35cm}
	\caption{\emph{Midpoint noise level $\mu$} for the example of LeNet-5/CIFAR-10/BN and globally injected additive noise.}
	\label{fig:method:acc_v_std_global_LeNet-BN-CIFAR10}
	\vspace{-0.4cm}
\end{figure*}

For all datasets, we train a given network and noise combination until it converges.
As a general pattern, one can observe how the network accuracy $a$ degrades with increased noise $\sigma$ (referring to the standard deviation of the injected Gaussian noise) until the network stops learning entirely.
Figure~\ref{fig:method:acc_v_std_global_LeNet-BN-CIFAR10} shows the resulting correlation on the example of LeNet-5 with BatchNorm trained on CIFAR-10.

\subsection{Quantifying robustness}

To quantify the robustness as a key performance metric, we define the \emph{midpoint noise level} $\mu$, also called robustness, where the injected noise is such that the network achieves half of its maximum accuracy, precisely $\delta a=(a_{max}-a_{min})/2$, with $a_{max}$ and $a_{min}$ being the maximally and minimally achieved accuracy, respectively.
Under normal circumstances $a_{min}$ equates to the random prediction accuracy for any given dataset, 
which is $1/K$ for $K$ classes in a classification task.
In this work, it is about $a_{min}=8.3\%$ for the GSC dataset and $a_{min}=10\%$ for all other datasets.
Equation~\ref{eq:logistic_CDF} then describes a scaled and shifted logistic function which is fitted to the observed data,
\begin{equation} \label{eq:logistic_CDF}
	F(\sigma; \mu, s, \delta a, a_{min}) = \frac{2}{1+e^{(\sigma-\mu)/s}} \cdot \delta a + a_{min}
\end{equation}
with $\mu$ as specified before,
$s$ the curve's slope, and $\delta a$ and $a_{min}$ being the curves scale and shift factor along the y-axis, illustrated in Figure~\ref{fig:method:acc_v_std_global_LeNet-BN-CIFAR10}.
Fitting a function to the data additionally gives us the ability to elegantly link data points with uncertainty information. 
Such uncertainties are a natural result of fluctuations during training.
Thus the fit also returns an error estimate, which can be used to assess the reliability of the obtained result:
\begin{equation} \label{eq:logistic_fit}
	\mu = \ \underset{\mu, s, \delta a, a_{min}}{\arg\min} \  \norm*{\frac{F(\sigma; \mu, s, \delta a, a_{min}) - y(\sigma)}{\Delta y(\sigma)}}^2, \forall \sigma
\end{equation}
where $y(\sigma)$ and $\Delta y(\sigma)$ refer to the observed accuracy and uncertainty, respectively.
Please note that $\mu$ here is both: Result and parameter to be determined by the fitting function. This metric follows the intuition that a larger $\mu$ corresponds to more robustness against noise.

It should be noted, that this is roughly equivalent to simply finding the datapoint closest to the threshold, where the investigated network achieves half of the maximum possible accuracy. 
However, the \emph{midpoint noise level} metric can furthermore statistically quantify how certain an observation is.
\begin{confidential}
Refer to Figure~\ref{fig:apdx:mul:double_fit_2} in the appendix for an example where a simple threshold would fail.
\end{confidential}

\begin{table}%
	\centering
	\small
	\caption{Robustness $\mu$ to globally injected noise (a larger $\mu$ corresponds to more robustness) for different neural architectures and datasets, with and without BatchNorm (BN).}
	\begin{tabular}{c|c|cc|cc}
		&                 & \multicolumn{2}{c|}{\textbf{Additive Noise}} & \multicolumn{2}{c}{\textbf{Multipl. Noise}} \\
		\textbf{Dataset} & \textbf{Model}  & \textbf{w/o BN} & \textbf{w/t BN} & \textbf{w/o BN} & \textbf{w/t BN} \\ \hline
		\multirow{2}{*}{\textbf{MNIST}}     & MLP    & 1.69  & 1.57 & 0.942 & 0.952 \\
		& LeNet-5 & 1.34  & 1.61 & 0.946 & 0.966 \\ \hline
		\multirow{2}{*}{\textbf{FashionMNIST}}   & MLP    & 2.76  & 2.42 & 0.801 & 0.802 \\
		& LeNet-5 & 2.10  & 2.36 & 0.828 & 0.838 \\ \hline
		\multirow{2}{*}{\textbf{SVHN}}   & MLP    & 1.94  & 1.76 & 0.623 & 0.620 \\
		& LeNet-5 & 1.65  & 1.68 & 0.627 & 0.652 \\ \hline
		\multirow{2}{*}{\textbf{CIFAR-10}}   & MLP    & 2.89  & 2.11 & 0.655 & 0.642 \\
		& LeNet-5 & 1.62  & 1.95 & 0.625 & 0.629 \\ \hline
		\multirow{2}{*}{\textbf{GSC}}      & MLP    & 5.83  & 2.06 & 0.670 & 0.667 \\
		& CNN-S  & 7.89  & 3.73 & 0.916 & 0.772 
	\end{tabular}
	\label{tab:method:global_noise}
\end{table}

\subsection{Robustness results for global noise}
Table~\ref{tab:method:global_noise} reports robustness for all datasets, models and regularization methods. %
We do not inject additional noise at the BatchNorm layers, to keep the amount of overall injected noise constant between experiments.
For globally injected noise all models show a higher robustness against additive noise compared to multiplicative noise.
Still, particular questions, for instance if BatchNorm is helpful or not, remain inconclusive from these results.

%% file: content/4-add-noise.tex
\section{Additive noise}
\label{sec:add_noise}

To further understand the layer-wise behavior under noise, we now consider additive noise in combination with the \emph{Walking Noise} methodology.
Considering constant noise, one would expect that a good learning procedure improves the signal-to-noise ratio by learning larger weights, which was also observed by~\cite{NoisyMachines}.
A vast body of previous work is reporting particular sensitivity in the input and output layers with regard to quantization.
As this is a lossy compression technique and  also introduces a kind of noise, we expect similar results here.

\begin{figure*}%
	\centering
	\includegraphics[width=0.99\textwidth]{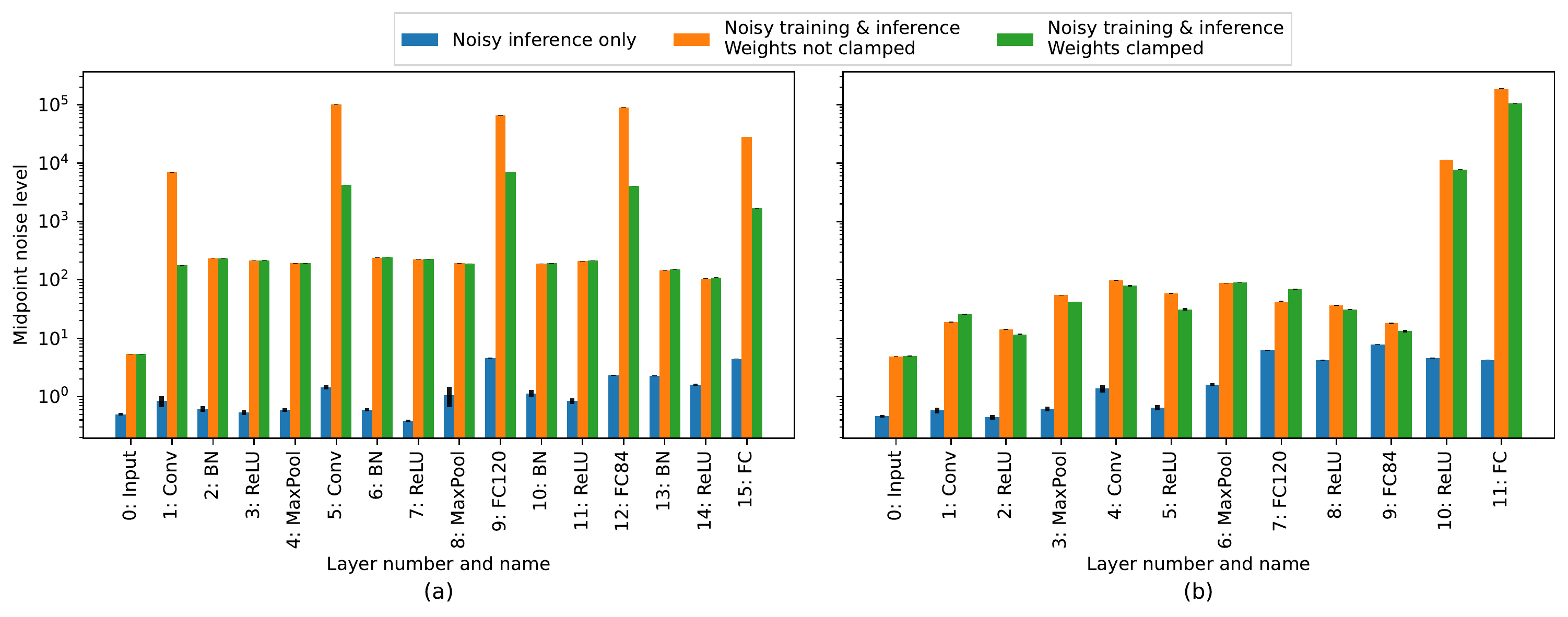}
	\vspace{-0.4cm}
	\caption{Layer-specific robustness to additive noise, for LeNet-5/CIFAR-10, with (a) and without (b) BatchNorm.}
	\label{fig:add:agg_LeNet-both-CIFAR10}
	\vspace{-0.4cm}
\end{figure*}

To investigate these expectations, we optionally prevent the weights from increasing without limit, by clamping them ($w \in [-1, 1]$) during training. %
We report the layer-specific robustness (\emph{midpoint noise level}) as key metric in Figure~\ref{fig:add:agg_LeNet-both-CIFAR10} for training with and without BatchNorm.
These results (LeNet-5/CIFAR-10) are representative of what we qualitatively observed for other architectures and datasets%
\begin{confidential}
	\footnote{Please refer to \Cref{fig:apdx:add:agg_GSC,fig:apdx:add:agg_ct} in the appendix for corresponding results.}%
\end{confidential}
.
We report a baseline (blue) without weight clamping and no noise exposure during training, results for noise applied during training (orange), and results for noise during training with clamped weights (green).

\subsection{Impact of batch normalization}
From the results illustrated in Figure~\ref{fig:add:agg_LeNet-both-CIFAR10}(a) it is immediately apparent that the network becomes significantly more robust when applying noise also during training and that especially layers with learnable parameters (convolutions and fully-connected layers) are much more robust.
Most of the \emph{midpoint noise level} values of non-learnable layers appear to plateau around a fixed value.
These plateaus are directly caused by the BatchNorm layers, which effectively nullify any absolute increase in the activation values, by scaling them to an approximate standard normal distribution $\mathcal{N}(0,1)$.
However, for layers with learnable parameters the network can introduce significantly larger activation values directly after the layer, which leads to notably higher robustness. 
While, the first and last layers show less robustness, which is in-line with our second expectation. 

These observations are much less visible when BatchNorm is disabled (Figure~\ref{fig:add:agg_LeNet-both-CIFAR10}(b)).
Here one can still clearly see that training with noise injection is more robust than without, but the results are less regular and the overall robustness is reduced.
Excluding layers 10 and 11, one can still see a trend from low robustness at the input to higher robustness at internal layers, back to less robustness at the output. 
Interestingly, the last two layers show exceptional robustness compared to the others,
which can be explained by a combination of two effects: Compared to previous layers, the overall injected noise is less, since the output layers are small, leading to less noisy elements. 
Additionally, layer 9 has much more parameters (84x120) compared to the output activations (10), therefore, it can create large activation values, even when using overall small weights. 

\begin{table}[htbp]
	\small
	\centering
	\caption{Ratio between clamped and non-clamped average weight magnitudes.}
	\begin{tabular}{c|c|c|c|c|c}
		\textbf{Layer ID} & \textbf{1} & \textbf{2} & \textbf{3} & \textbf{4} & \textbf{5} \\ \hline
		\textbf{Layer Type} & Conv & Conv & FC & FC & FC  \\ 
		\textbf{with BatchNorm}   & 51      & 30      & 13    & 24    & 18           \\
		\textbf{without BatchNorm}   & 1.7     & 0.9    & 1.1   & 0.6   & 1.1   
	\end{tabular}
	\label{tab:method:weight_mag}
	\vspace{-1.1cm}
\end{table}

\subsection{Impact of weight magnitude}
In Figure~\ref{fig:add:agg_LeNet-both-CIFAR10}(a) a clear contrast between clamped and non-clamped weights becomes apparent, as unclamped weights result in higher robustness.
This suggests that layers with learnable parameters learn larger weights as noise is increased.
Surprisingly, disabling BatchNorm results in almost no difference between training with clamped weights and without, which leads to the hypothesis that a model architecture without BatchNorm is less likely to learn arbitrarily large weights.
By analyzing the weight magnitudes of a network, with injected noise closest to \emph{midpoint noise level}, we can confirm this.
The average weight magnitudes with and without BatchNorm are compared in Table~\ref{tab:method:weight_mag}, by considering the ratio between clamped and non-clamped weights magnitudes.
These results reproduce expectations and verify results found by \textit{Noisy Machines}~\cite{NoisyMachines}, that in such cases weights grow tremendously in magnitude, and in addition reveal the central role of BatchNorm for this effect.

%% file: content/5-mul-noise.tex
\begin{figure*}%
	\centering
	\includegraphics[width=0.69\textwidth]{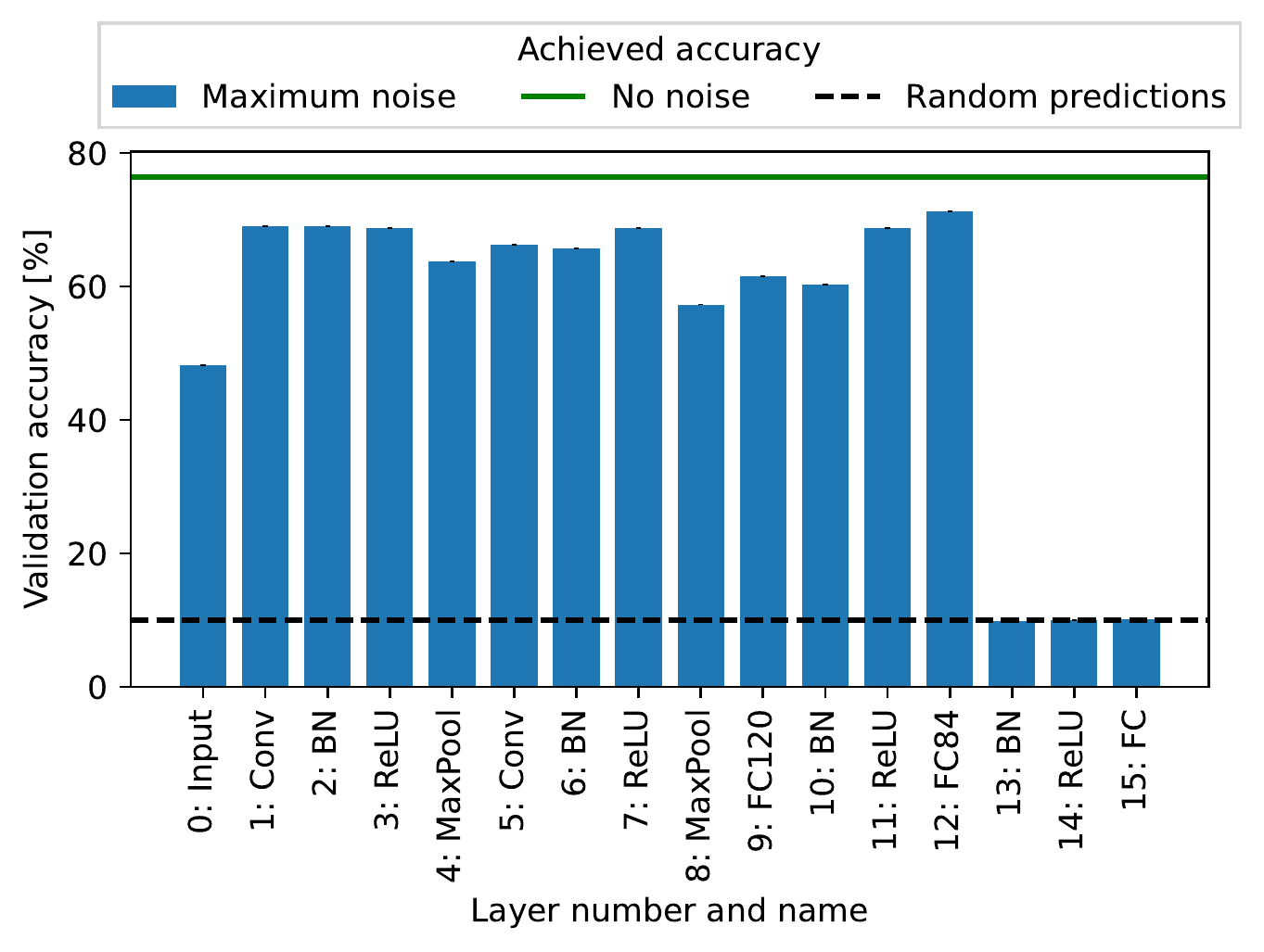}
	\vspace{-0.35cm}
	\caption{Noise tolerance per layer: validation accuracy for LeNet-5/CIFAR-10 with BatchNorm, when injecting large amounts of noise multiplicatively ($\sigma=10^{10}$) at each layer. Only the last layers drop to random prediction accuracy.}
	\label{fig:mul:agg_LeNet-BN-CIFAR10}
	\vspace{-0.4cm}
\end{figure*}
\section{Multiplicative noise}
\label{sec:mul_noise}
Initial assumptions were that the results for the multiplicatively injected noise would be similar to additively injected noise.
The only difference in expectation was that the network would not learn larger weights as this would not influence the signal-to-noise ratio.
However, our experiments with multiplicative \emph{Walking Noise} showed that in model architectures with BatchNorm 
some layers withstood noise injected 
with a standard deviation of up to $10^{10}$, without ever dropping more than $10\%$ points in accuracy on the CIFAR-10 task.
This was especially surprising as
previously no architecture could withstand noise injections larger than $1$ on a global scale without
significant accuracy degradations (Table \ref{tab:method:global_noise}).
Furthermore, this notably deviates from the previous considerations on $a_{min}$, as accuracy does not deteriorate to $1/K$ for $K$ classes.
Thus, for curve fitting we here make use of $a_{min}$ instead of $\mu$ (see Eq. \ref{eq:logistic_CDF}).
Figure~\ref{fig:mul:agg_LeNet-BN-CIFAR10} shows the drop in accuracy per layer for LeNet-5 on CIFAR-10, when BatchNorm is enabled and a large amount of noise is applied.

\begin{figure*}[ht!]%
	\centering
	\includegraphics[width=0.7\textwidth]{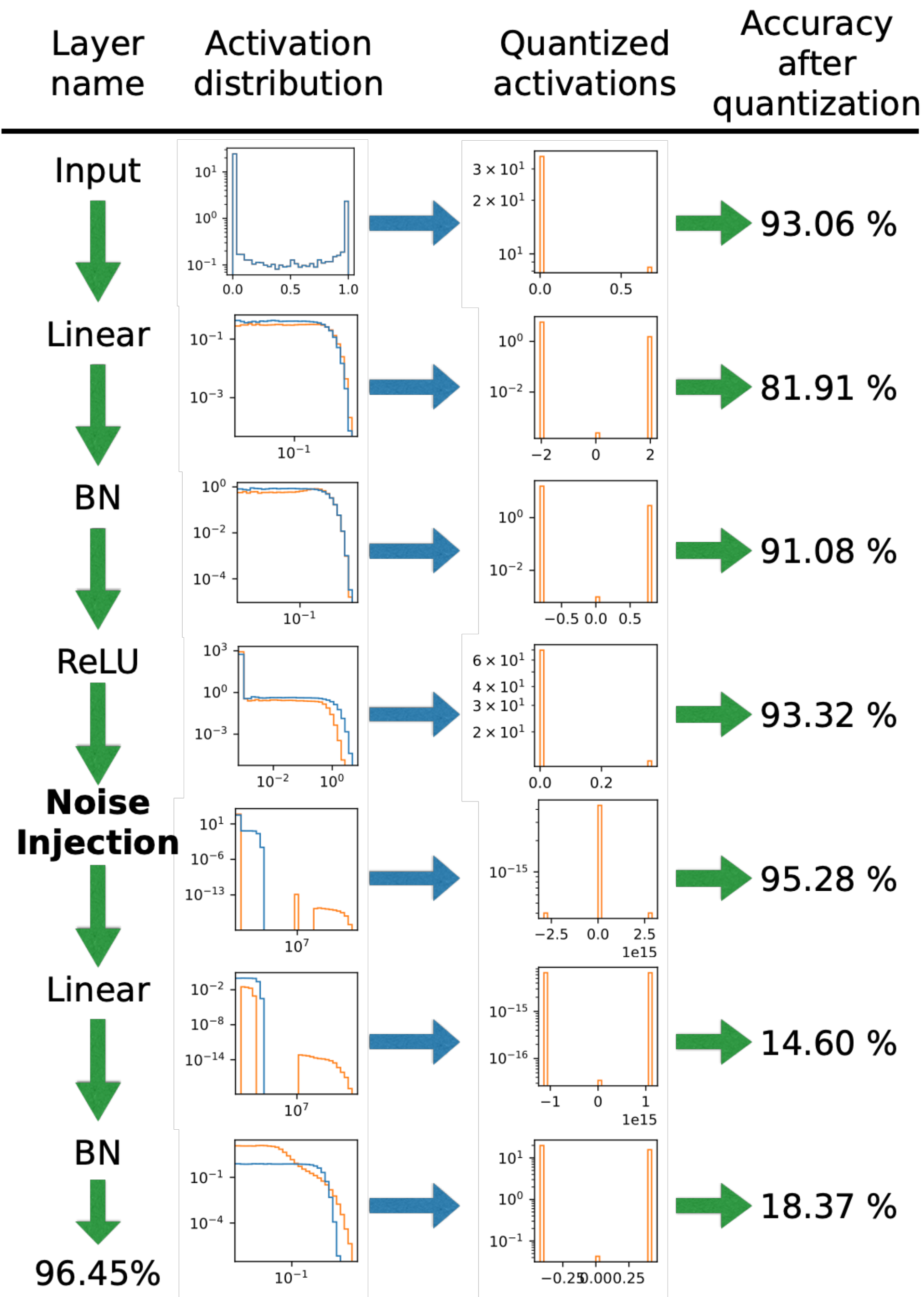}
	\vspace{-0.35cm}
	\caption{Visualization of \emph{self-binarization} of an MLP on MNIST for multiplicative \emph{Walking Noise}. The left histograms show the distribution of activation values with noise (orange) and without (blue) after the layer is executed:
		Logarithmic and linear x-axis scales are used to highlight variations between the two networks. 
		On the right side the distribution of activation values with achieved accuracy is shown when applying a simple threshold based quantization to the respective layer.}
	\label{fig:mul:self-bin}
	\vspace{-0.6cm}
\end{figure*}

To better understand this apparently very scalable robustness to multiplicative noise, we investigate how the activations behave during inference.
The left column of Figure~\ref{fig:mul:self-bin} illustrates the activations for each layer as histograms, 
with large amounts of multiplicative noise injected.
For most cases the distributions with and without noise are of similar shape, but after the noise injection layer,
the noisy activation values split into two peaks of a bimodal distribution.
One mode clusters around zero, while the other clusters just below the standard deviation of the injected noise. 
This suggests that the network is able to distribute its activation values into two distinct clusters.

\subsection{Self-binarization of model activations}

We hypothesize that the network is able to recover information solely from these two peaks, without the need for any exact value within the peaks themselves, since the values within have been effectively randomized by noise injection. 
The network would then effectively encode a learned binary representation and afterwards decode it again as a ternary distribution, since the noise acts symmetrically.
To investigate if this is what we were observing, we introduced an additional simple threshold-based quantization step to alter the in-flight activations during the computation.
To be exact, we selectively quantize the activations of a layer into two or three peaks: the values centered around zero were set to be exactly zero and the values around the second/third peak were set to their average.
The right side of Figure~\ref{fig:mul:self-bin} shows the activation histograms and accuracies after this quantization. %
Since this operation does not preserve the shape of the peaks, but reduces them to one value, 
the accuracy should drop significantly when the network is not able to tolerate this binarization.
However, as accuracy is only marginally impacted at the noise injection, 
these experiments support our hypothesis that the network has learned a binary representation by itself. 
As a result, 
the distinction between zero and any large number is one way, if not the primary one, in which a model can reliably circumvent multiplicative noise.

\begin{figure*}%
	\centering
	\includegraphics[width=0.7\columnwidth]{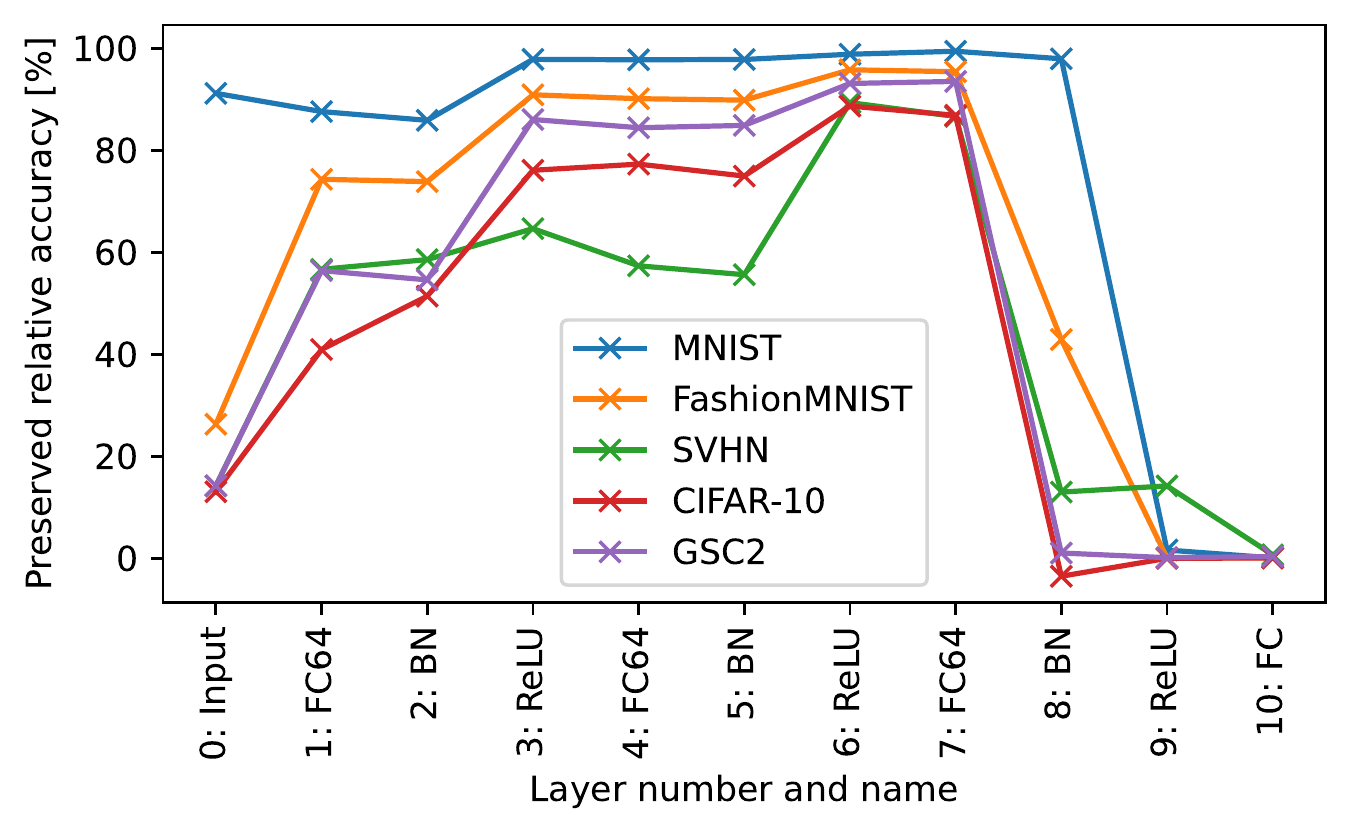}
	\vspace{-0.35cm}
	\caption{Accuracy preservation for MLP with BatchNorm and various datasets, noise injected multiplicatively.}
	\label{fig:mul:MLP_BN_compare_datasets}
	\vspace{-0.4cm}
\end{figure*}

Since this form of robustness appeared across all datasets and models we are able to compare how the same model architecture performs on different datasets, on a per-layer basis.
Figure~\ref{fig:mul:MLP_BN_compare_datasets} illustrates the \emph{preserved relative accuracy} for each layer of the MLP for different datasets.
The metric is calculated by taking the ratio of the model accuracy at no noise and the accuracy at maximum noise, thus making it possible to compare different datasets, even if the baseline accuracy is dissimilar. 
It should be noted that accuracy in and of itself is a highly non-linear and network-dependent metric, so while general trends are a helpful insight, the absolute values are to be taken with a grain of salt.
One can see that on MNIST the MLP is able to preserve accuracy by far the best, pointing towards the simplicity of the data.
The most complex datasets on the other hand appear to be CIFAR-10 and SVHN.
Interestingly, this offers a point of reference on how complex the GSC dataset is in comparison, even though the task is from a different domain.
One can also observe the general trend that, as the noise injection happens deeper in the model architecture, the networks ability to counteract the noise improves, until layer eight or nine, where the accuracy sharply drops to random accuracy. 
We assume that this final drop is due to the network not being able to properly decode any 
self-learned binarization without a BatchNorm layer behind the noise injection point.

These results do extend to more complex model architectures, such as convolutional ones, showing that self-binarization appears to be possible independent of architecture.
In particular more complex model architectures appear to be able to make even better use of this effect%
\begin{confidential}
	, see Figure~\ref{fig:apdx:mul:compare_dataset} in the appendix%
\end{confidential}
.

In contrast, \emph{globally injected noise} also affects the last layers, which are necessary to decode the binary representation.
Which restricts the robustness through self-binarization in this case.
In order to maintain robustness, the last layers would have to be shielded in some way from noise.

\subsection{The impact of batch normalization}

While BatchNorm appears to be essential for this extreme robustness against noise,
there are indications that even without it some self-binarization is possible. 
Without this regularization we observed that the accuracy against noise injection curves (compare Figure~\ref{fig:method:acc_v_std_global_LeNet-BN-CIFAR10}) would no longer follow a single logistic function, but would first drop to a plateau, before dropping down to random prediction accuracy%
\begin{confidential}
	, as illustrated in Figure~\ref{fig:apdx:mul:double_fit_2} in the appendix%
\end{confidential}
.
This behavior can be described by two stacked logistic functions, suggesting that two independent effects are at play.
We assume that the plateau is caused by intermittent self-binarization, before the network can no longer compensate for the increased activation magnitude and the accuracy drops down completely.

%% file: content/5b-comb-noise.tex
\section{Mixed noise}
\label{sec:comb_noise}
Since the self-learned robustness strategies for additive and multiplicative noise vary greatly, the learning behavior in a mixed noise setting proves interesting, particularly whether the network learns to self-binarize, accompanied by extreme robustness, in an environment with a significant amount of additive noise.
\begin{figure*}[tbh]
	\centering
	\includegraphics[width=0.90\textwidth]{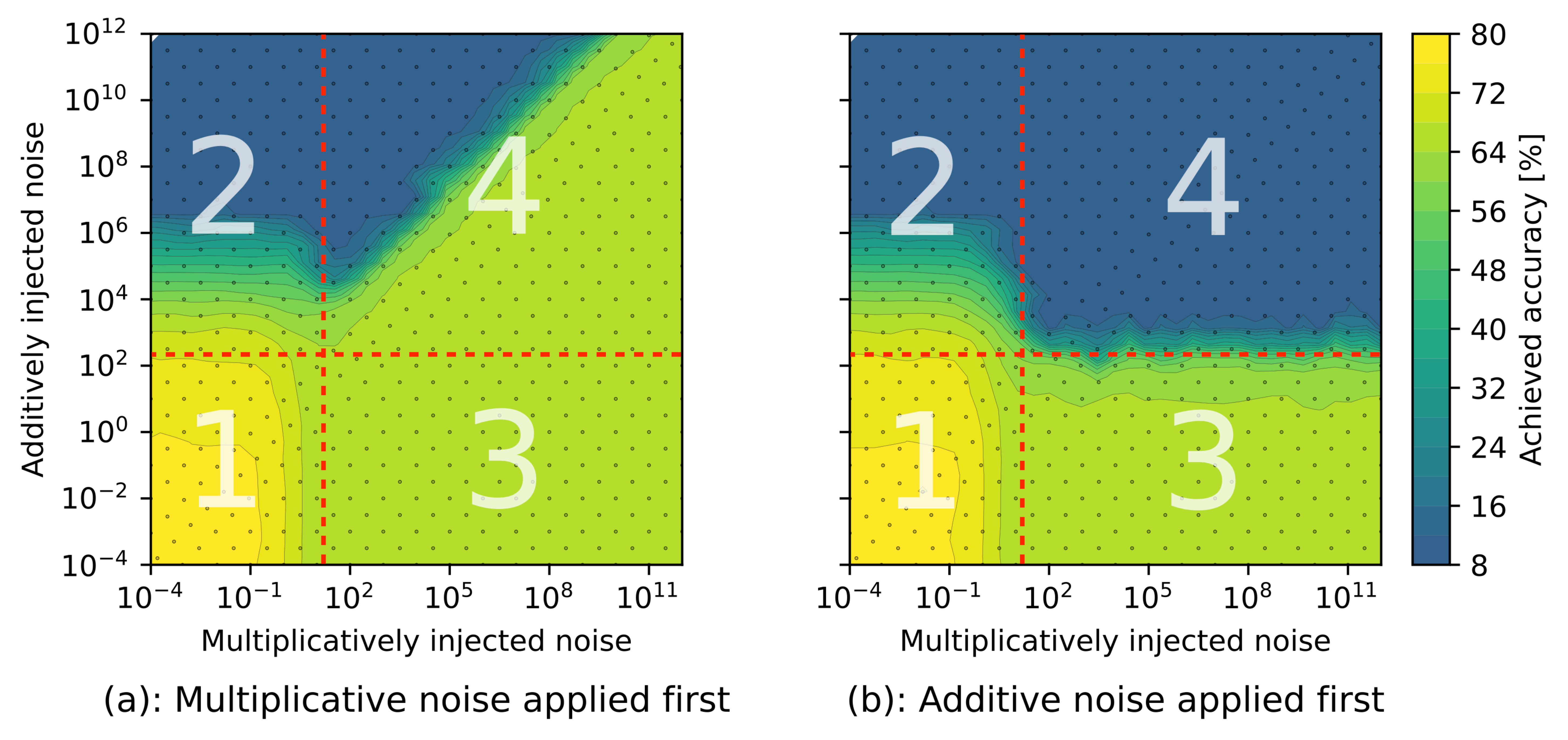}
	\caption{Accuracy for training with multiplicative and additive noise at layer 5 (Conv) of LeNet-5 trained on CIFAR-10. The black dots indicate points of measurement.}
	\label{fig:comb:contour_LeNet-BN-CIFAR10_both}
\end{figure*}

In real-world analog matrix-multiply accelerator for instance multiple sources of noise exist and it depends on the specific hardware whether multiplicative or additive noise components occur first in the computational order.
We therefore evaluate both the \emph{multiplicative-first} and \emph{additive-first} case,
\begin{equation}
\begin{split}
\text{multiplicative-first: } y(x) &= \big(x \cdot \mathcal{N}(1,\sigma^{2}_\text{mul})\big) + \mathcal{N}(0,\sigma^{2}_\text{add}) \\
\text{additive-first: } y(x) &= \big(x + \mathcal{N}(0,\sigma^{2}_\text{add})\big) \cdot \mathcal{N}(1,\sigma^{2}_\text{mul})
\end{split}
\end{equation}
for value $x$ and noise sampled from a Gaussian distribution $\mathcal{N}(\mu,\sigma^{2})$ with standard deviations for the multiplicative $\sigma_\text{mul}$ and additive $\sigma_\text{add}$ noise components.

\Cref{fig:comb:contour_LeNet-BN-CIFAR10_both} illustrates the experimental results with both computational orders.
Here one can see that for cases where one type of noise is dominant (regimes 1 to 3) previous results are reproduced (\Cref{sec:add_noise,sec:mul_noise}).
Further more, we find that when both noise components are strong (regime 4), then the final network performance highly depends on the order of operations.
In particular we observe that the network can still learn to self-binarize, if multiplicative noise is applied first, thus allowing for extreme robustness.
Interestingly this effect persists, even when additive noise of up to one order of magnitude more is injected afterwards.
For practical accelerator designs this means that a hardware-in-the-loop trained network may tolerate additive noise sources up to very high degrees, if the additive noise is not substantially larger and time-wise appears after the multiplicative noise.
\begin{confidential}
	Section~\ref{sec:apdx:mixed_noise} in the appendix discusses the associated experiments, findings and implications in detail.
\end{confidential}

%% file: content/6b-using_WN.tex
\section{Making use of Walking Noise results}
\label{sec:using_walking_noise}

With the \emph{Walking Noise} methodology, we now have a measure of a neural architecture's robustness at per-layer resolution. 
It can characterize either the pure inference dynamics or also provide information about the training dynamics when noise is present.

The potential applications of such information are manifold.
Essentially, we see \emph{Walking Noise} as a method on the interfaces in between neural architecture design, processor design, and bridging methods such as model compression as well as countermeasures against unwanted model behavior in general, such as adversarial settings.
Most obviously the method is interesting for investigating analog neural network accelerators, which is the overarching goal of this work.
However, in practice the results from \emph{Walking Noise} experiments can be representative of any perturbation, which can reasonably be interpreted as a form of either multiplicative or additive noise. Other selected examples include:
\begin{itemize}
	\item 	Understanding the implications of different noise forms on prediction quality. This might furthermore include the design space of analog accelerators (different trade-offs among processing speed and noise), as well as searching for robust architectures in the presence of different noise forms and amounts (neural architecture search).
	\item 	Informed compression techniques such as quantization and pruning. Knowledge about the sensitivity of a layer to noise is anticipated to correlate with its redundancy, thus is expected to be a valuable information when deciding about compression parameters such as bit width, data type or sparsity. In contrast to experiments for a specific compression method, which may have only a few discrete configurations, the sensitivity to noise, with its continuous spectrum, promises to be more general.
	\item 	In adversarial settings, \emph{Walking Noise} can be used to either increase the robustness to input perturbations, or to quantify the benefits of existing adversarial training methods with regard to noise tolerance. Possible connections to neural architecture search are highly anticipated. 
	\item 	Ensembles respectively redundancy by multi-execution is a natural approach to improve noise robustness in various settings, and \emph{Walking Noise} can be a guide to how much redundancy is needed in which layer, leading to more tailored and resource-efficient methods.
\end{itemize}

\begin{table}[htbp]
	\centering
	\small
	\caption{Walking Noise guiding multi-execution to improve accuracy.}
	\begin{tabular}{c|c|c|>{\tiny}l>{\tiny}l|cc}
		&                 & & \multicolumn{2}{c|}{\textbf{Executions per layer}} & \multicolumn{2}{c}{\textbf{Accuracy}} \\
		\textbf{Dataset} & \textbf{Model}  & \textbf{BN} & {\small \textbf{Uniform}} & {\small \textbf{Guided}} & \textbf{Uniform} & \textbf{Guided} \\ \hline
		MNIST & MLP & with 		& \{2,2,2,...\} & \{6,1,3,3,2,1,2,1,1,1,1\} &  $71.4\pm0.7$\,\% & $\mathbf{80.6\pm0.5}$\,\% \\
		MNIST & MLP & without 		& \{2,2,2,...\} & \{6,1,3,3,2,1,2,1,1,1,1\} &  $68.2\pm1.3$\,\% & $\mathbf{87.1\pm0.7}$\,\% \\
		CIFAR-10 & MLP & with 		& \{2,2,2,...\} & \{1,1,3,4,2,2,3,2,1,2,1\} &  $41.7\pm0.5$\,\% & $\mathbf{43.7\pm0.5}$\,\% \\
		CIFAR-10 & MLP & without 	& \{2,2,2,...\} & \{4,1,1,1,1,1,2,5\} &  $38.9\pm0.6$\,\% & $\mathbf{43.4\pm0.7}$\,\% \\
		CIFAR-10 & LeNet-5 & with 	& \{2,2,2,...\} & \{3,2,3,3,3,1,3,4,2,1,1,2,1,1,1,1\} &  $58.5\pm1.4$\,\% & $\mathbf{61.1\pm1.0}$\,\% \\
		CIFAR-10 & LeNet-5 & without & \{2,2,2,...\} & \{4,3,4,3,1,3,1,1,1,1,1,1\} &  $57.2\pm1.8$\,\% & $\mathbf{62.4\pm1.3}$\,\% \\
	\end{tabular}
	\label{tab:repetitions}
\end{table}

As a first simple application we test multi-execution in noisy environments as it is closely related to analog accelerators. 
Here we test how well a network can be inferred when allowed to execute layers $N$ times, thus reducing the overall noise perturbation by $1/\sqrt{N}$. %
As a baseline we distribute the number of executions uniformly over all layers of a network. Using the data gained by our \emph{Walking Noise} experiments, we redistribute the total amount of executions such that more sensitive layers get executed more often than others.
For such an application, we consider the reciprocal value of midpoint noise to obtain the number of repetitions $n_i$ for each layer $i$, normalized to the total number of repeated executions $n_t$.
Applying sum-conserving rounding, we then obtain the desired number of repetitions per layer, while $n_t$ is conserved and $n_i$ are valid natural numbers.

To demonstrate its effectiveness we select a challenging setting with significant globally injected additive noise, equivalent to the obtained global midpoint noise%
\begin{confidential}
	,  from Table~\ref{tab:method:global_noise_inf} in the appendix%
\end{confidential}
.
The results are shown in Table~\ref{tab:repetitions} and illustrate that in a very noisy setting and even without retraining, the \emph{Walking Noise} layer-wise sensitivity information can be a helpful guidance for a more effective countermeasure.

In general, we observe that neural architectures are becoming more complex in depth, operator diversity and architecture itself, complicating the understanding of noise sensitivity.
Here \emph{Walking Noise} can help to reason about sensitivity in an automatically generated, fine-granular manner.

%% file: content/6-summary.tex
\section{Summary}
\label{sec:summary}

This work considers energy-efficient but noisy analog computations to provide an understanding of the fundamental behavior of neural architectures with regard to robustness against such noise.
As in reality there are various forms of noise in analog computations, we present an abstract model of noisy computations that injects additive, multiplicative and mixed Gaussian noise at the activations between layers. We introduce the metric \emph{midpoint noise level} to assess the robustness of an architecture against noisy computations, and increase its granularity to the different components of a neural architecture by employing a method called \emph{Walking Noise}.

Our results include general information on robustness to absolute noise levels and the impact of BatchNorm.
We discover fundamentally different behavior of additive and multiplicative noise. 
For additive noise we observe increased growth in the magnitude of weight parameters which improves the signal-to-noise ratio, and confirm that components closer to the input and output of an architecture are more sensitive to noise. 
In contrast, for multiplicative noise we gain the insight that models self-binarize during training, which results in extreme robustness to noise.
Notably, in both cases BatchNorm plays a central role, which adds more aspects to the ongoing discussions about BatchNorm~\cite{NEURIPS2018_905056c1,8953671}.
When combining both noise types we find that for small perturbations the network conforms to previous findings, but for large perturbations the order in which the noise is injected is crucial.
This provides insights for designers of analog processors as well as practitioners concerned with the deployment of neural architectures on such analog processors.
We demonstrate the use of \emph{Walking Noise} on the example of tailored multi-execution to minimize the overhead of this simple robustness method.

We understand this work as a first step towards an understanding of more complex model architectures, such as residual and attention-based networks.
Additionally, by revealing a networks internal robustness characteristics, \emph{Walking Noise} provides an alternative perspective to evaluate and optimize existing robustness methods for noisy computations.
Ultimately, we are interested in finding the most robust model architecture and most effective hardening method for a given type of noise, as well as guiding processor design with regard to tolerable amounts and types of noise.

%% file: content/8-Appendix.tex
\section*{Appendix}
\subsection{Methodology}

\begin{figure}[tbh]
	\centering
	\includegraphics[width=0.99\columnwidth]{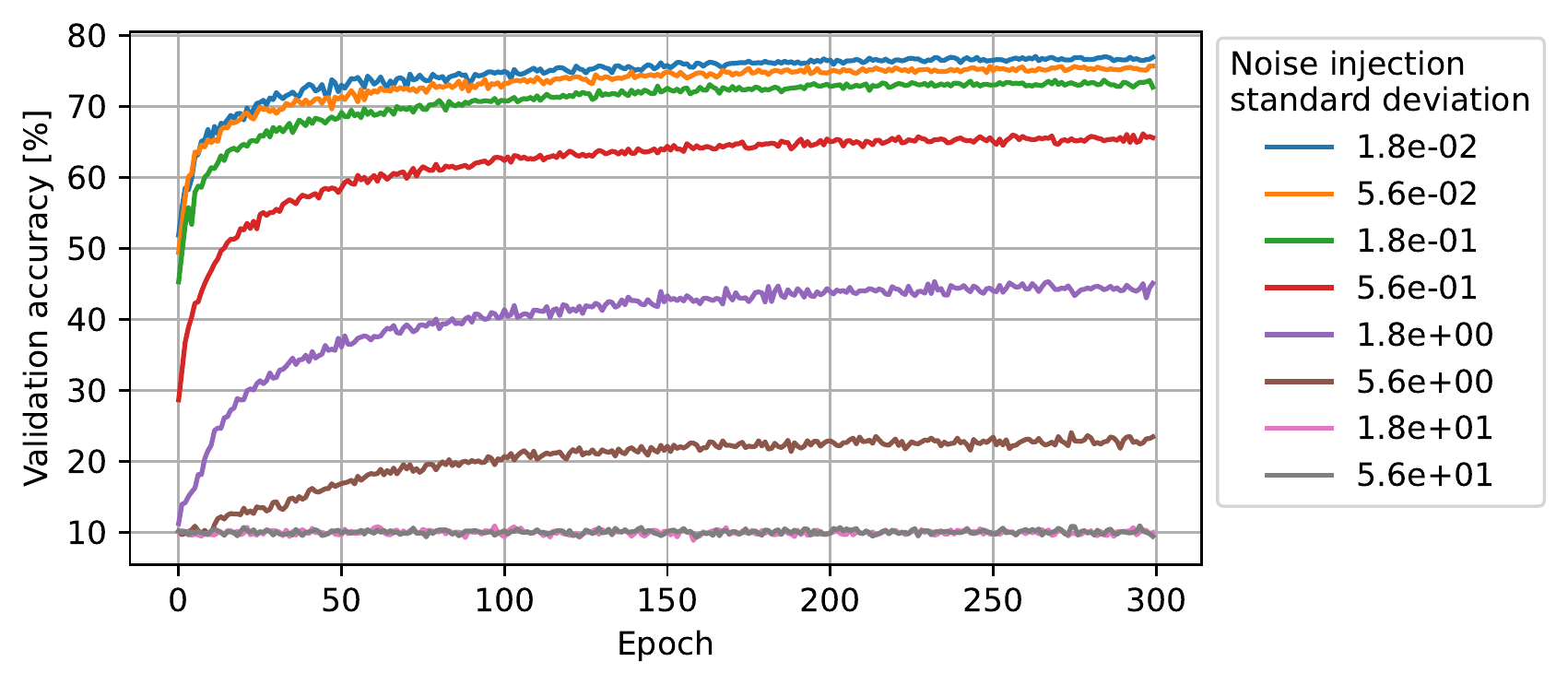}
	\caption{Accuracy for LeNet-5/CIFAR-10 with training epochs for globally injected additive noise. One can observe how an increasing amount of noise deteriorates the validation accuracy until it arrives to random prediction accuracy.}
	\label{fig:apdx:method:acc_v_epoch_global_LeNet-BN-CIFAR10}
\end{figure}

To extend on how a given network is trained under noise injection, Figure~ \ref{fig:apdx:method:acc_v_epoch_global_LeNet-BN-CIFAR10} highlights the training process for LeNet-5 on the CIFAR-10 task, trained with noise injection.
While the network converges after some epochs to non-noise accuracy for small additive noise values, an increasing amount of noise deteriorates the validation accuracy until it stops learning entirely and the accuracy becomes equivalent to random prediction accuracy.

\subsection{Global noise, without training}
Equivalent to Table~\ref{tab:method:global_noise} the midpoint noise can be found for all networks, without \emph{noisy training}.
In this case the network is first trained without noise, and noise is only injected during inference.
Without \emph{noisy training} the models can tolerate significantly less global noise, which is evident in the much smaller midpoint noise values in Table~\ref{tab:method:global_noise_inf} compared to Table~\ref{tab:method:global_noise}.
\begin{table}[htbp]
	\centering
	\small
	\begin{tabular}{c|c|cc|cc}
		&                 & \multicolumn{2}{c|}{\textbf{Add. Noise}} & \multicolumn{2}{c}{\textbf{Multipl. Noise}} \\
		\textbf{Dataset} & \textbf{Model}  & \textbf{BN} & \textbf{w/o BN} & \textbf{BN} & \textbf{w/o BN} \\ \hline
		\multirow{2}{*}{\textbf{ MNIST }} &  MLP  & 0.35 & 0.47 & 0.45 & 0.55 \\ 
		&  LeNet  & 0.24 & 0.61 & 0.39 & 0.55 \\ \hline
		\textbf{Fashion-} &  MLP  & 0.30 & 1.19 & 0.25 & 0.40 \\ 
		\textbf{MNIST}&  LeNet  & 0.09 & 0.17 & 0.23 & 0.35 \\ \hline
		\multirow{2}{*}{\textbf{ SVHN }} &  MLP  & 0.23 & 1.41 & 0.23 & 0.31 \\ 
		&  LeNet  & 0.15 & 0.30 & 0.30 & 0.34 \\ \hline
		\multirow{2}{*}{\textbf{ CIFAR10 }} &  MLP  & 0.33 & 1.63 & 0.30 & 0.37 \\ 
		&  LeNet  & 0.20 & 0.25 & 0.22 & 0.28 \\ \hline
		\multirow{2}{*}{\textbf{ GSC2 }} &  MLP  & 0.18 & 1.47 & 0.27 & 0.30 \\ 
		&  CNN-S  & 0.24 & 1.85 & 0.34 & 0.48
		
	\end{tabular}
	\caption{Robustness $\mu$ to globally injected noise, when injecting noise during inference only.}
	\label{tab:method:global_noise_inf}
\end{table}

\subsection{Multiplicative noise}
\subsubsection{Walking Noise without BatchNorm:}
\begin{figure}[tbh]
	\centering
	\includegraphics[width=0.99\columnwidth]{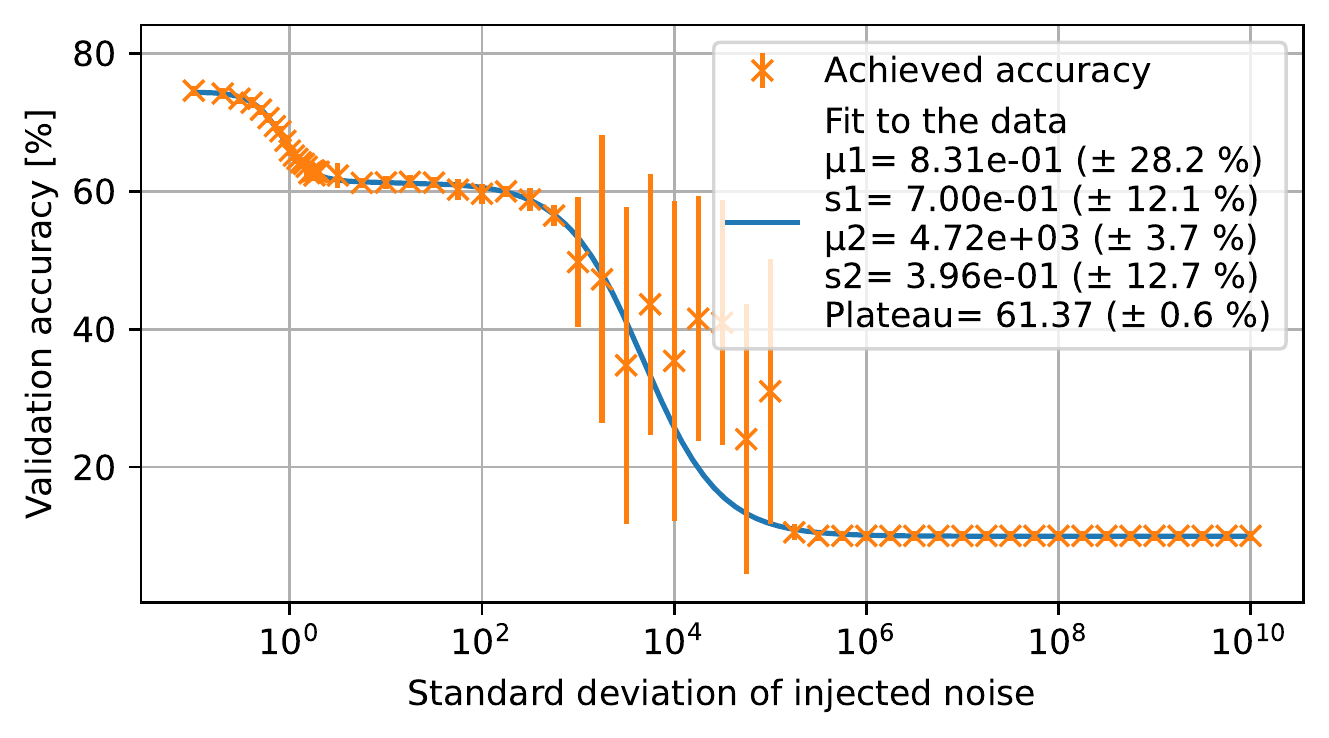}
	\caption{Accuracy for LeNet-5/CIFAR-10 with a fit of two stacked logistic functions, when injecting noise multiplicatively at the third layer in the network without BatchNorm.}
	\label{fig:apdx:mul:double_fit_2}
\end{figure}

Extending on the effects of multiplicative noise, Figure~\ref{fig:apdx:mul:double_fit_2} shows how validation accuracy degradation is intermittently paused when no batch normalization is used with LeNet-5 on the CIFAR-10 task.
This plot in particular shows how the data can be described by two stacked logistic functions, highlighting that there are likely two distinct effects at play.
The example also demonstrates the usefulness of measuring the uncertainty of data points by repeating experiments, since otherwise it would be impossible to reliably fit a function to the parts with large error bars.

\subsubsection{Preserved accuracy with BatchNorm for other model architectures:}
\begin{figure*}[tbh] 
	\subfloat[LeNet-5\label{fig:apdx:mul:LeNet_BN_compare_datasets}]{%
		\includegraphics[width=0.49\columnwidth,height=0.35\columnwidth]{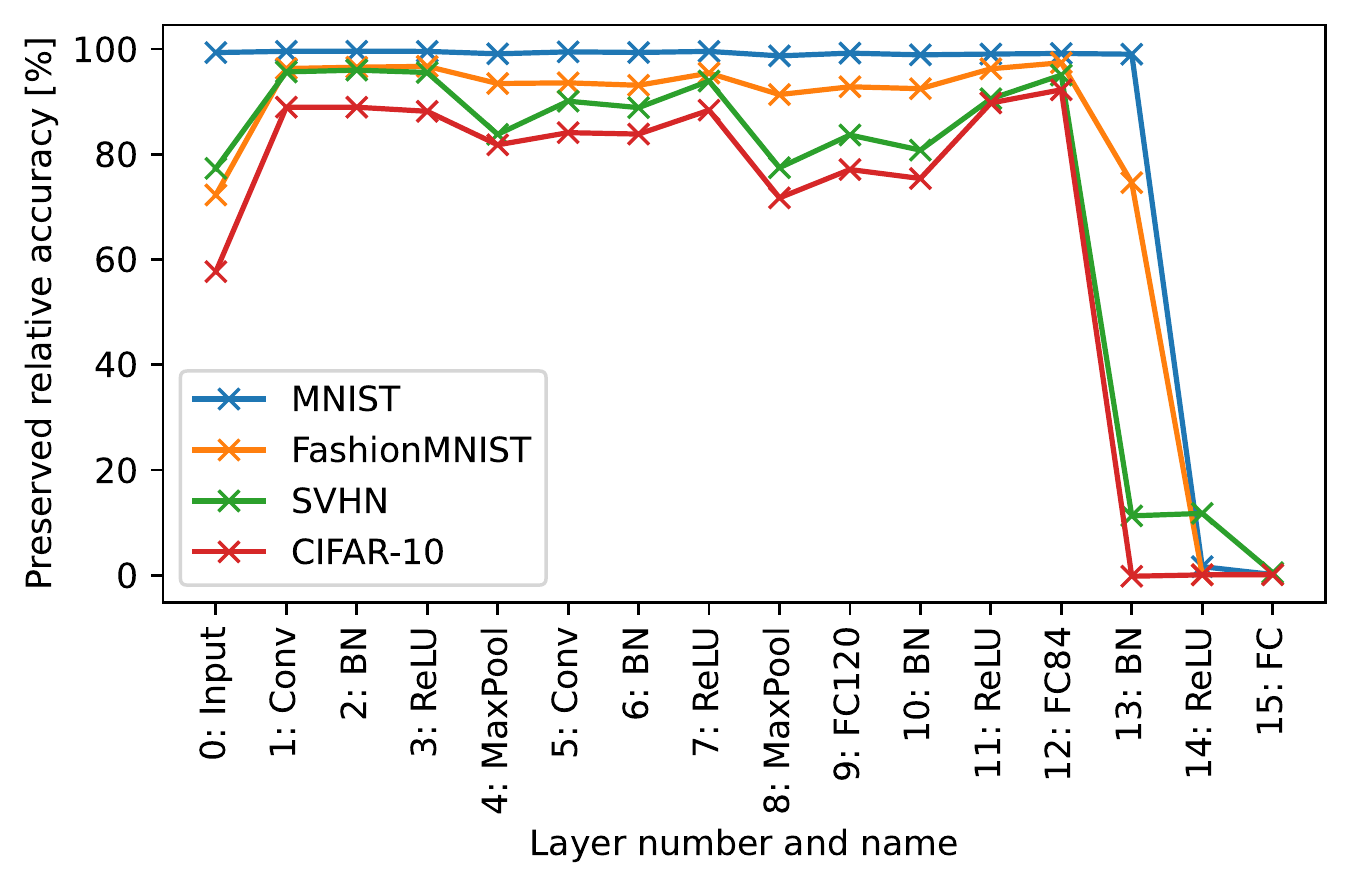}}
	\subfloat[CNN-S\label{fig:apdx:mul:CNN_BN_compare_datasets}]{%
		\includegraphics[width=0.49\columnwidth,height=0.35\columnwidth]{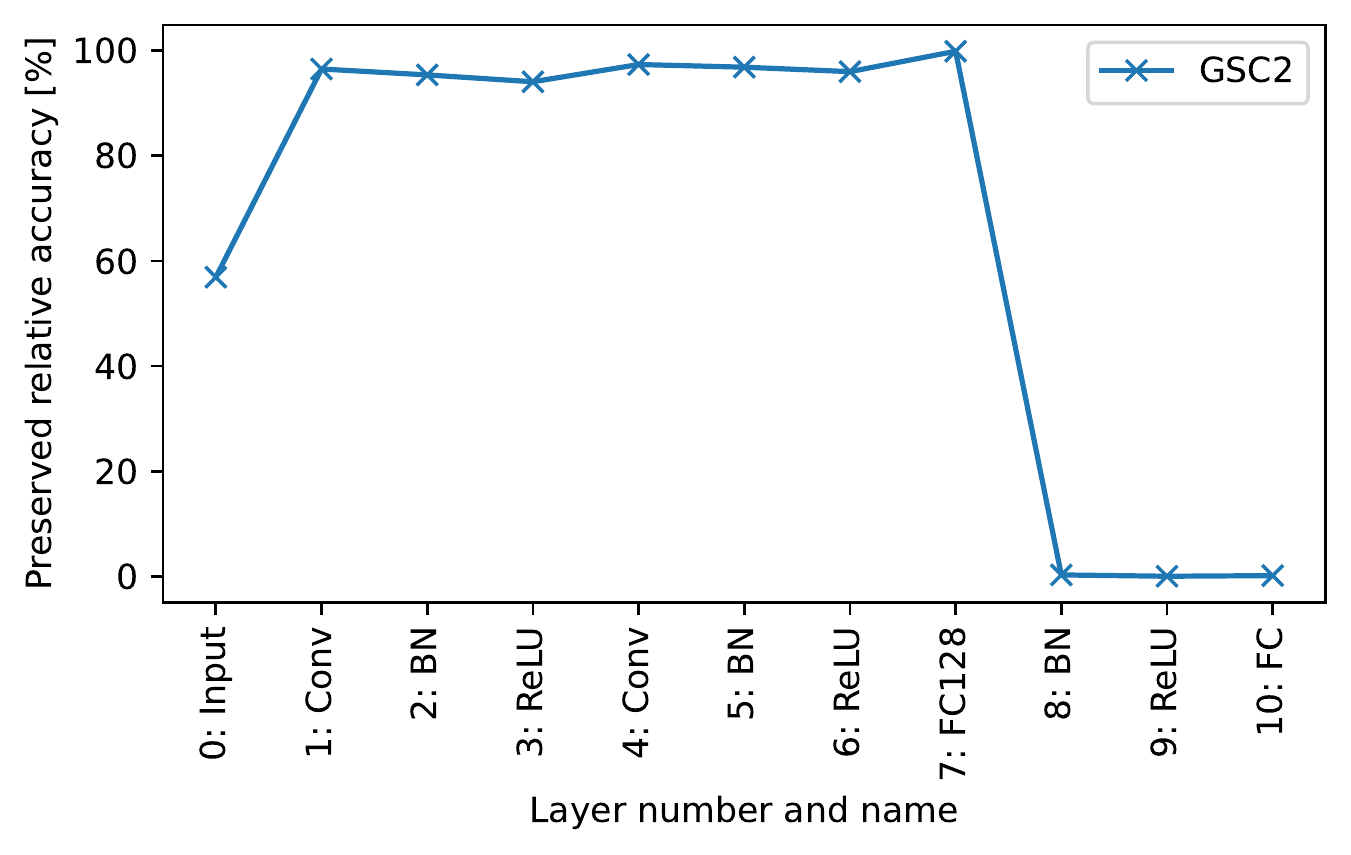}}
	\caption{Accuracy preservation for LeNet-5 and CNN-S on different datasets, when injecting noise multiplicatively at a given point in the network with BatchNorm.}
	\label{fig:apdx:mul:compare_dataset}
\end{figure*}

To highlight that model accuracy can be preserved for multiplicative noise, given BatchNorm regularization, \Cref{fig:apdx:mul:compare_dataset} shows the preserved relative accuracy for the other model architectures investigated in this work.
It is particularly notable that the accuracy is preserved to a certain degree for all architectures and datasets, excluding the last two to three layers.

\subsection{Mixed Moise}
\subsubsection{Self-binarization and robustness under mixed noise:}
\label{sec:apdx:mixed_noise}

In this part of the appendix we'd like to expand on the results from \Cref{sec:comb_noise}.
To investigate the self-binarization under mixed noise we selected an example layer --- Number 5 of LeNet-5 with BatchNorm on CIFAR10 --- which had previously shown good robustness to additive noise and self-binarization capabilities.
We then explored a wide range of combinations for multiplicative and additive noise injection strengths and observed the networks validation accuracy under \emph{noisy training}.
Figure~\ref{fig:comb:contour_LeNet-BN-CIFAR10_both} illustrates these accuracy heatmaps, where the black dots indicate independent training and evaluation points, while the contour lines and thus colors are linearly interpolated.
In particular, four regimes can be distinguished:
\begin{itemize}
	\item Regime 1: Both noise types are low and the network trains almost to maximum accuracy.
	\item Regime 2: The injected noise is dominated by additive noise and the network trains as explained in section~\ref{sec:add_noise}.
	\item Regime 3: The injected noise is dominated by the multiplicative component and the model trains as discussed in section~\ref{sec:mul_noise}.
	\item Regime 4: Both additive and multiplicative noise are high, here the result highly depends on the order of operations. %
\end{itemize}

For the \emph{multiplicative-first} case, one can discover in total four different regimes of network behavior.
Within the plateau of high accuracy, for low multiplicative and additive noise, the network trains without issue to full accuracy, as perturbations from the noise are minimal (regime 1). 
Notably, the network is more robust to additive noise, in terms of absolute noise injection values, independent of the order of operations.
When keeping either the additive or multiplicative component of the mixed noise low, the high-accuracy regime smoothly transitions into the behavior already observed for either noise type exclusively in sections \ref{sec:add_noise} and \ref{sec:mul_noise}: For low multiplicative noise with increasing additive noise we observe a gradual transition to random prediction accuracy (regime 2); and for low additive noise with increasing multiplicative noise the network gradually goes over to self-binarization (regime 3).
However, previously unseen results appear at the intersection of high multiplicative and additive noise (regime 4).
Here we observe that the learned self-binarization can overcome significant additive noise injection.
More precisely it appears that first indications of a loss of accuracy for the self-binarized models appear, when the additive noise is already larger than the multiplicative one by almost an order of magnitude. (Compare the first diagonal contour line in the fourth regime and at high accuracy in Figure~\ref{fig:comb:contour_LeNet-BN-CIFAR10_both}.)
After this, with more additive noise the accuracy drops rapidly, which is expected, as the quantized values created by the self-binarization should now be firmly below the values of the additive noise, thus having no effect on the accuracy of the network.

When switching the order of noise injection to \emph{additive-first} the situation changes notably, illustrated in Figure~\ref{fig:comb:contour_LeNet-BN-CIFAR10_both}(b). 
While most observations remain as they were, the previously discovered self-binarization, occurring while both noise types are large, vanishes.
Here, the accuracy instead drops to random prediction accuracy.
This illustrates, that the additive noise term removes all information from the activations, before any self-binarization effects can manifest.

One of the key insights is that for small contributions of any one noise component, the training remains unperturbed and follows the observations from sections \ref{sec:add_noise} and \ref{sec:mul_noise}, independent on the order of injection.
The picture however changes when both noise components are relatively large and of similar size.
When first applying additive noise, \emph{noisy training} cannot compensate for the disastrous effect of noise.
However, when multiplicative noise is injected first, the network can learn to self-binarize.
Here we observe that high accuracy results can still be achieved even when the additive component is larger than the multiplicative one, up to about a factor of 10.
Interestingly, through the learned self-binarization a network with additionally injected multiplicative noise upfront can be orders of magnitude more robust than a network with additive noise only, although the total amount of injected noise is larger.
This could be especially interesting in situations where the additive noise in a system is prohibitively strong, and increasing the multiplicative component might be a simple but non-intuitive option to improve the overall network robustness.

\subsection{Additive noise}
\begin{figure*} 
	\centering
	\subfloat[MLP on GSC.\label{fig:apdx:add:agg_MLP-GSC2}]{%
		\includegraphics[width=0.90\textwidth]{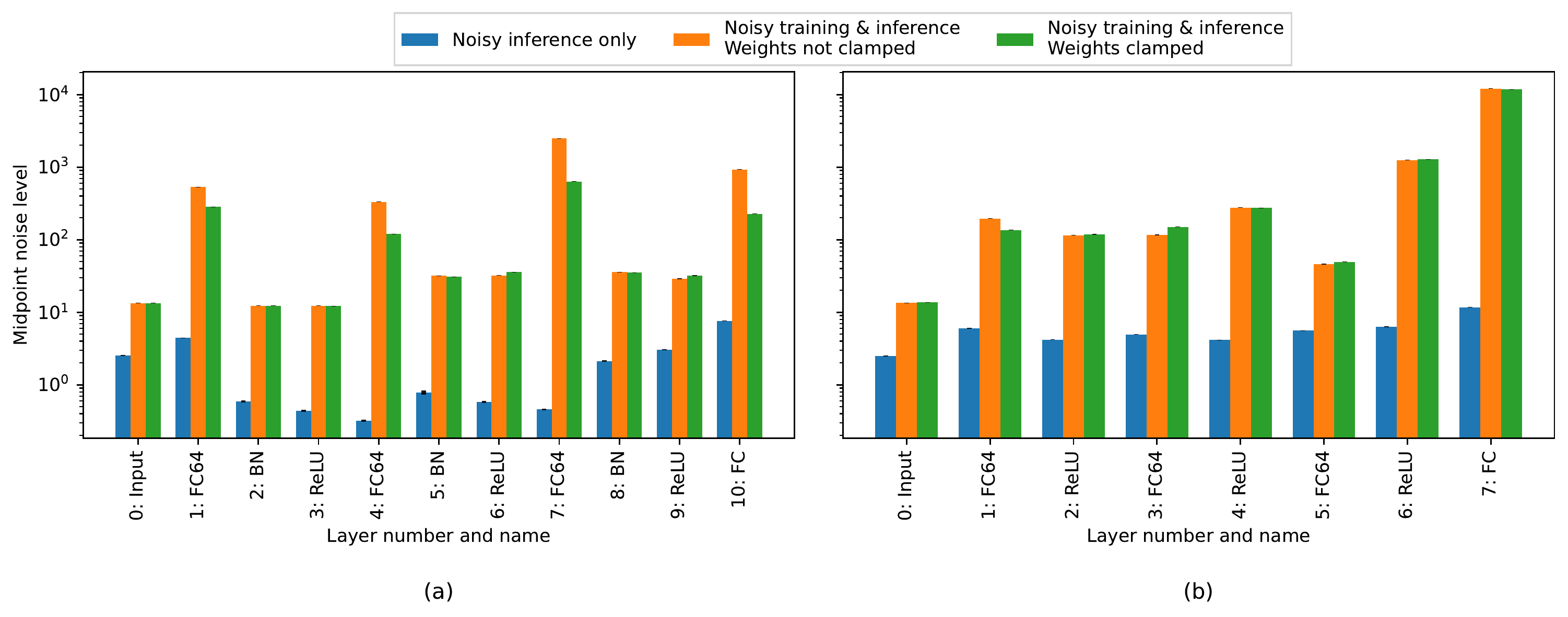}}
	\\
	\subfloat[CNN-S on GSC.\label{fig:apdx:add:agg_CNN-GSC2}]{%
		\includegraphics[width=0.90\textwidth]{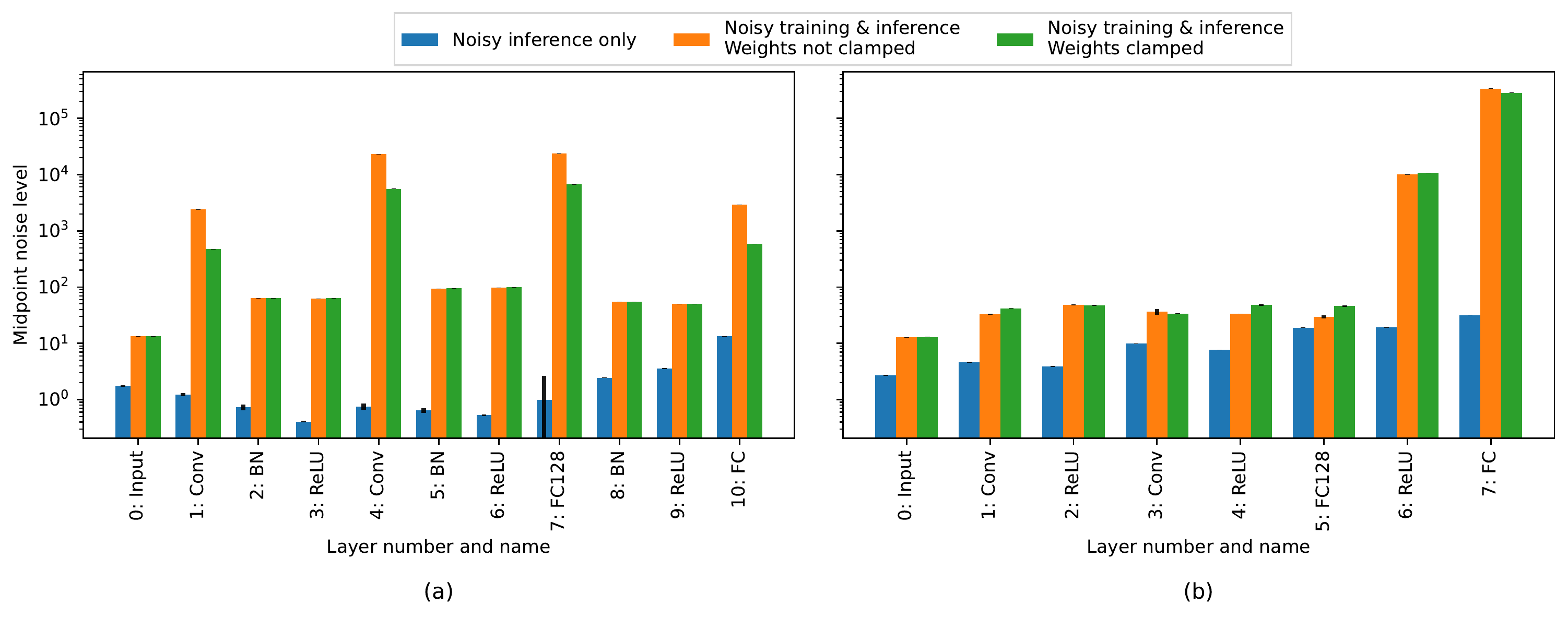}}
	\\
	\caption{Midpoint noise level for various model architectures on Google Speech Commands based on Walking Noise. The x axis shows layer number and name. Figures (a) to the left are with BatchNorm, Figures (b) to the right are without BatchNorm.}
	\label{fig:apdx:add:agg_GSC}
\end{figure*}

\begin{figure*} 
	\centering
	\subfloat[MLP on MNIST.\label{fig:apdx:add:agg_MLP-MNIST}]{%
		\includegraphics[width=0.90\textwidth]{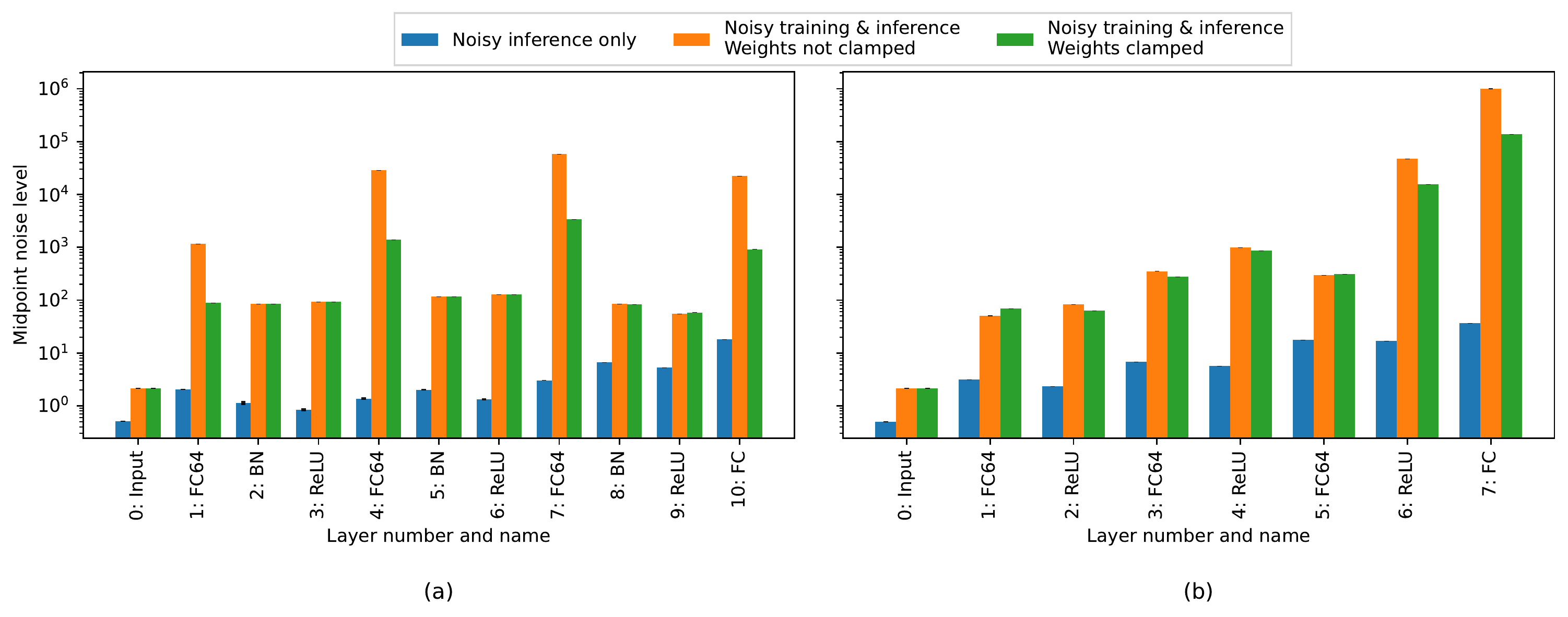}}
	\\
	\subfloat[LeNet-5 on MNIST.\label{fig:apdx:add:agg_LeNet-MNIST}]{%
		\includegraphics[width=0.90\textwidth]{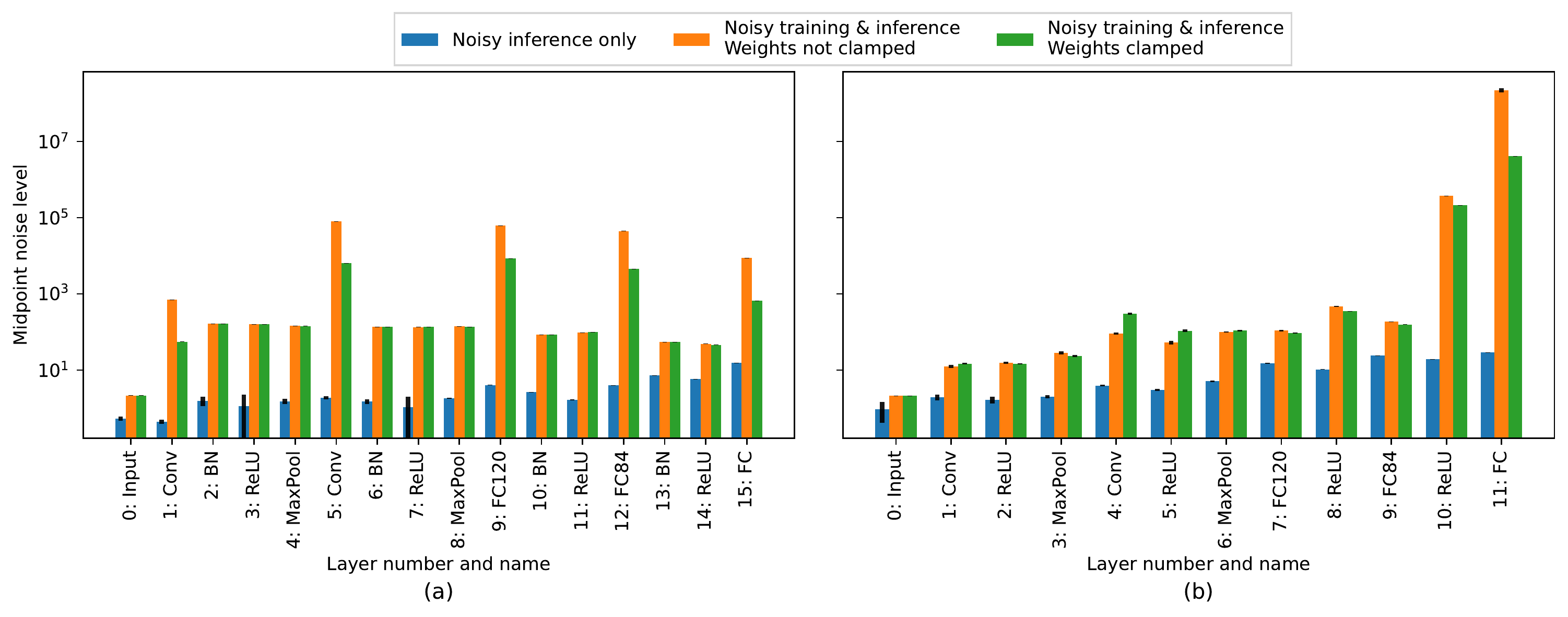}}
	\\
	\subfloat[MLP on CIFAR-10.\label{fig:apdx:add:agg_MLP-CIFAR10}]{%
		\includegraphics[width=0.90\textwidth]{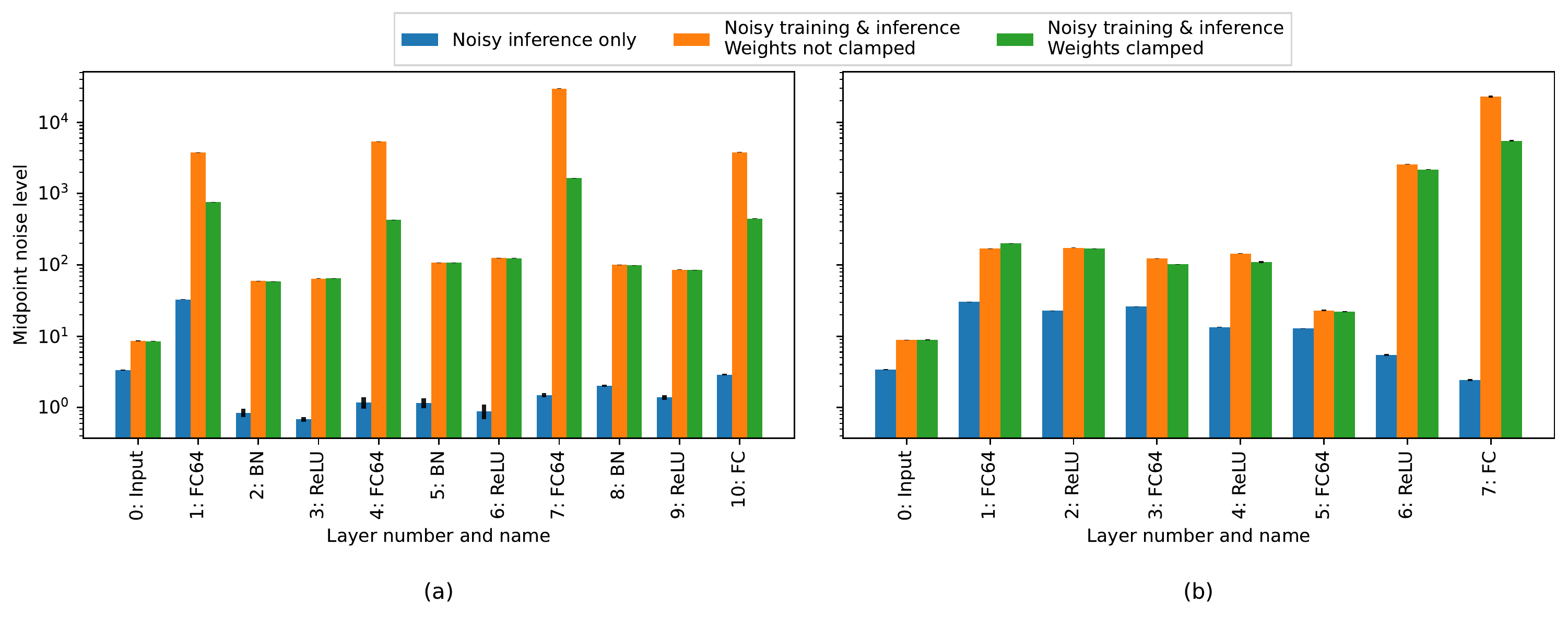}}
	\caption{Midpoint noise level for various model architectures and image classification datasets based on Walking Noise. The x axis shows layer number and name. Figures (a) to the left are with BatchNorm, Figures (b) to the right are without BatchNorm.}
	\label{fig:apdx:add:agg_ct}
\end{figure*}

\paragraph{Other model architectures and datasets:}
To highlight that the injection of additive noise with the Walking Noise methodology behaves similar on all investigated model architectures and datasets, \Cref{fig:apdx:add:agg_GSC} for Google Speech Commands and \Cref{fig:apdx:add:agg_ct} for MNIST and CIFAR-10 show similar noise tolerance patterns as illustrated in \Cref{fig:add:agg_LeNet-both-CIFAR10}.
In particular these Figures show results for the MLP, LeNet-5 and CNN-S architectures.

